\documentclass[journal]{IEEEtran}
\usepackage{cite}
\usepackage[utf8]{inputenc} 
\usepackage[T1]{fontenc}    
\usepackage{hyperref}       
\usepackage{url}            
\usepackage{booktabs}       
\usepackage{amsfonts}       
\usepackage{nicefrac}       
\usepackage{microtype}      
\usepackage{lipsum}
\usepackage{algorithm}
\usepackage{algorithmic}
\usepackage{graphicx}
\usepackage{multirow}
\usepackage{tikz}
\usetikzlibrary{positioning, fit, arrows.meta, shapes, calc}

\tikzset{
	every path/.style={->, >=stealth, very thick, rounded corners},
	state_box/.style={
		rectangle, rounded corners,
		draw=black, fill=gray!20, thick,
		minimum height=1em, inner sep=6pt, text centered
	}
}

\begin{document}

\title{Effect of backdoor attacks over the complexity of the latent space distribution}

\author{Henry~Chac\'on,
        and~Peyman~Najafirad
\thanks{H. Chac\'on is with the Department of Management Science and Statistics, University of Texas at San Antonio, San Antonio,
TX, 78249 USA, e-mail: henry.chacon@utsa.edu.}
\thanks{P. Najafirad is with the Department of Information Systems and Cyber Security, University of Texas at San Antonio, San Antonio,
	TX, 78249 USA, e-mail: peyman.najafirad@utsa.edu.}
}


\markboth{}%
{Chac\'on and Najafirad: Bare Demo of IEEEtran.cls for IEEE Journals}

\maketitle

\begin{abstract}
	The input space complexity determines the model's capabilities to extract their knowledge and translate the space of attributes into a function which is assumed in general, as a concatenation of non-linear functions between layers. In the presence of backdoor attacks, the space complexity changes, and induces similarities between classes that directly affect the model's training. As a consequence, the model tends to overfit the input set. In this research, we suggest the D-vine Copula Auto-Encoder (VCAE) as a tool to estimate the latent space distribution under the presence of backdoor triggers. Since no assumptions are made on the distribution estimation, like in Variational Autoencoders (VAE). It is possible to observe the backdoor stamp in non-attacked categories randomly generated. We exhibit the differences between a clean model (baseline) and the attacked one (backdoor) in a pairwise representation of the distribution. The idea is to illustrate the dependency structure change in the input space induced by backdoor features. Finally, we quantify the entropy's changes and the Kullback–Leibler divergence between models. In our results, we found the entropy in the latent space increases by around 27\% due to the backdoor trigger added to the input.\footnote{Code available at:
		
		 \href{https://github.com/henrychacon/Backdoor_attacks/tree/main/D-Vine_copula_auto_encoder}{https://github.com/henrychacon/Backdoor\_attacks/tree/main/D-Vine\_copula\_auto\_encoder}} 
\end{abstract}


\IEEEpeerreviewmaketitle

\section{Introduction}

Advances on Machine and Deep Learning models have positioned those techniques in an accelerated momentum process, gaining attention in the scientific community due to the diversity of applications \cite{bendre2020learning}. Pre-trained models and methods such as transfer learning have provided access to practitioners willing to reduce training time and computational cost required for models with a large number of parameters and big data sizes. However, it opens new possibilities of Cyber data attacks in terms of fairness and poisoning \cite{silva2020opportunities}. In most cases, final users can not validate if the model is free off adversarial or backdoor attacks. A situation reflected in a survey published in 2020 by \cite{kumar2020adversarial} over 28 organizations from different sectors. They found that 25 out of those surveyed companies ``\textit{do not have the right tools in place to secure their ML systems and are explicitly looking for guidance}''. Only two of the surveyed companies reported they developed their ML models from scratch. 

\begin{figure}
	\centering
	\begin{tikzpicture}[node distance=4cm]
		\tikzstyle{terminal}=[rounded rectangle, minimum size=6mm, thick, draw=black, align=center,text width=42pt, fill=black!5,
		font=\scriptsize\sffamily]
		\tikzstyle{nonterminal}=[rectangle, minimum size=6mm, thick, draw=black, align=center, fill=black!25,
		font=\scriptsize\sffamily]
		\tikzstyle{nodeSamples} = [draw=black, fill=black!5, circle, minimum size=8mm]
		
		\filldraw[fill=blue!20, draw=black] (-0.1, -0.4) -- (-0.1, 0.4) -- (0.1, 0.4) -- (0.1, -0.4) -- cycle; 
		
		\filldraw[fill=green!20!white, draw=black] (0.2, 0.4) -- (2, 0.7) -- (2, -0.7) -- (0.2, -0.4) -- cycle; 
		\filldraw[fill=green!20!white, draw=black] (-0.2, 0.4) -- (-2, 0.7) -- (-2, -0.7) -- (-0.2, -0.4) -- cycle; 
		
		\filldraw[fill=green!20!white, draw=black] (2.1, 0.7) -- (2.3, 0.7) -- (2.3, -0.7) -- (2.1, -0.7) -- cycle; 
		\filldraw[fill=green!20!white, draw=black] (-2.1, 0.7) -- (-2.3, 0.7) -- (-2.3, -0.7) -- (-2.1, -0.7) -- cycle; 
		
		\node[] at (-1.1, 0) {Encoder};
		\node[] at (1.1, 0) {Decoder};
		
		\node[] at (-2.7, 0) {$\mathbf{X}$};
		\node[] at (2.7, 0) {$\mathbf{X}'$};
		\node[] at (0, 0.7) {$\mathbf{h}$};

		\filldraw[fill=red!40, draw=black] (-0.1, -2.4) -- (-0.1, -1.6) -- (0.1, -1.6) -- (0.1, -2.4) -- cycle; 
		\filldraw[fill=green!20!white, draw=black] (0.2, -1.6) -- (2, -1.3) -- (2, -2.7) -- (0.2, -2.4) -- cycle; 
		\filldraw[fill=green!20!white, draw=black] (2.1, -1.3) -- (2.3, -1.3) -- (2.3, -2.7) -- (2.1, -2.7) -- cycle; 
		\filldraw[fill=yellow!20!white, draw=black] (-2, -2.5) -- (-2, -1.5) -- (-0.6, -1.5) -- (-0.6, -2.5) -- cycle; 
		
		\node[] at (-1.3, -1.8) {D-vine};
		\node[] at (-1.3, -2.2) {Copula};
		\node[] at (1.1, -2) {Decoder};
		\node[] at (2.7, -2) {$\mathbf{X}'$};
		\node[font=\ttfamily] at (0.2, -3.1) {Generative model};
		
		\draw[->] (-0.6, -2)--(-0.1, -2); 
		\draw[->] (0, -0.4)--(0, -1)--(-2.6, -1)--(-2.6, -2)--(-2.15, -2); 
		\draw[dashed] (-2.1,-2.8) rectangle (2.4, -1.2); 
		
	\end{tikzpicture}
	\caption{The generative D-vine autoencoder (VCAE) approach suggested by \cite{tagasovska2019copulas} and considered to evaluate the backdoor effect in the distribution of a lower representation of the data. After the AE is trained, the D-vine copula and the decoder are transformed into the generative model. $\mathbf{X}$ corresponds to the input set (attacked or clean), $\mathbf{h}$ the latent domain of the input, and $\mathbf{X}'$ the reconstructed output.}
	\label{fig:VCAE}	
\end{figure}
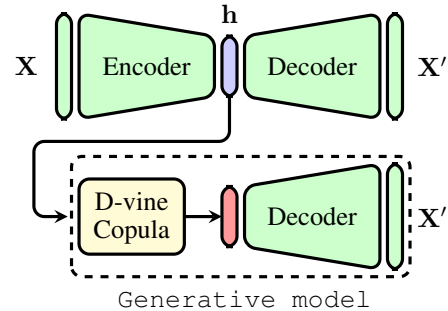


As artificial intelligence models have grown, the diversity of attacks has also evolved. According to \cite{vorobeychik2018adversarial}, ML attacks can be classified based on when the attack happens. When it happens before the training process, they are called poisoning or backdoor. If they are executed after the model is trained, they are known as inference time or adversarial attacks. The attacker's intention in the former is to produce a misclassification of the model denying access to one particular category or forcing the model to generate the same response if a trigger is present. On the other hand, adversarial attacks are intended to exploit the model's vulnerabilities by using corrupted inputs to gain access as if they were a valid entrance. 

Most of the applications for ML models found in the literature are related to image classification \cite{hao2020adversarial}. Several challenges can be mentioned in the image domain. One of the relevant is the number of dimensions and the relationship between pixels. As it is expected, their dependency structure is not linear. Therefore, most of the models developed to classify images are based on a concatenation of layers linked by non-linear functions. Other types of models are designed to translate the problem into a lower-dimensional representation of the input. This family is called generative models since their goal is to recreate new images based on the input space. See for instance \cite{kingma2013auto} where authors suggest the variational inference approach to estimate the latent space distribution of the autoencoder.  Another relevant reference in this area is presented by \cite{kos2018adversarial}. In their research, authors generate attacked images using variational auto-encoders (VAE) and VAE-GAN tested on a different data set, as an alternative adversarial generation. 

The limitation with the variational autoencoders approach is the rigid constraints imposed on the latent space to estimate its distribution. As it is pointed out by \cite{tagasovska2019copulas}. In this regard, we consider the Vine Copula Autoencoders (VCAE) proposed by these authors to overcome the restrictions associated with the backpropagation method. Our contribution is the study of the footprint effect induced in the latent space due to backdoor attacks to better estimate its distribution. The VCAE model considered is depicted in figure \ref{fig:VCAE}. The dimensional reduction of the input generated by the AE after training is used as input to the D-vine copula method. It estimates the latent space distribution by a process in pairs illustrated in the methodology section. To evaluate the backdoor effect, we generate two models, a baseline AE trained on clean data and another model with one category fully attacked with a backdoor trigger. The Kullback–Leibler divergence of the latent distributions between both models is computed to measure the changes between inputs induced by the backdoor attack. In order to track the backdoor and the D-vine method as a generative process to detect the attack. Clean data is used in an attacked model to produce the latent space distribution. A better representation of a situation where an attacked model is provided with a clean data set for testing. Although the data used do not have any backdoor feature, samples generated by the generative process depicted in figure \ref{fig:VCAE} exhibit the trigger shape, not only in the attacked category but also in other categories that share some similarities with the target input. This confirms the attack is present in the model's parameters as it is suggested by \cite{liu2017neural}. 

To the best of our knowledge, this is the first research that considers the VCAE to estimate the latent domain distribution under the backdoor attacks. This paper is constructed as follows: the first section is defines properly the motivations to use VCAE instead of VAE to estimate the distribution in the latent space. Next, a brief introduction of the D-vine copula approach is presented. Then, the methodology is described, followed by the last section illustrating experimental results in detail. Finally, conclusions and future work are discussed.


\section{Motivation and related work}

Deep neural networks (DNN) are defined as highly expressive architectures able to identify the semantic information presented in the input data set. In different fields, DNN has become the state of the art for image, speech, and text classification. However, a small perturbation in the training set can lead to an increment in the prediction error. As it was initially stated by \cite{Szegedy2013}, the effect of a specific perturbation is not a random artifact of learning, but it causes the same effect applied in a different subset of the input in another architecture. This property is formally defined as \textit{adversarial transferability} by \cite{papernot2017practical} using as hypothesis the dependency relationship between sequences of the same distribution and their implications on similar models.

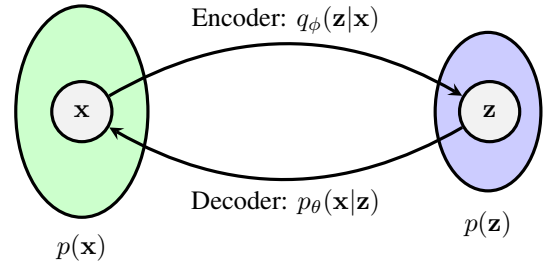
\begin{figure}
	\centering
	\begin{tikzpicture}[node distance=4cm]
		\tikzstyle{terminal}=[rounded rectangle, minimum size=6mm, thick, draw=black, align=center,text width=42pt, fill=black!5,
		font=\scriptsize\sffamily]
		\tikzstyle{nonterminal}=[rectangle, minimum size=6mm, thick, draw=black, align=center, fill=black!25,
		font=\scriptsize\sffamily]
		\tikzstyle{nodeSamples} = [draw=black, fill=black!5, circle, minimum size=8mm]
		
		\draw[x radius=25pt, y radius=40pt,  black,fill=green!20] (0,0) ellipse;
		\draw[x radius=20pt, y radius=30pt,  black,fill=blue!20] (5.4,0) ellipse;
		\draw[-] (0, -1.8) node {$p(\mathbf{x})$};
		\draw[-] (5.4, -1.5) node {$p(\mathbf{z})$};
		
		\node[nodeSamples] at (0, 0) (xsam) {$\mathbf{x}$};
		\node[right=of xsam, nodeSamples] at (1, 0) (zsam) {$\mathbf{z}$};
		
		\path[->] (xsam) edge[bend left] node[above]{Encoder: $q_{\phi}(\mathbf{z}|\mathbf{x})$}  (zsam);
		\path[->] (zsam) edge[bend left] node[below]{Decoder: $p_{\theta}(\mathbf{x}|\mathbf{z})$}  (xsam);
	\end{tikzpicture}
	\caption{Variational Autoencoder probabilistic graph representation. It is expected the latent space for $\mathbf{z} \in \mathbf{R}^n$ is smaller than the original input space $\mathbf{x} \in \mathbf{R}^m$ such as: $m > n$.}
	\label{fig:VAE}	
\end{figure}

Although the transferability property is fundamentally based on the distribution, most of its application is found on the training process and in the architecture design \cite{hao2020adversarial}. The reason is mainly associated with the DNN response under attacks. In the case of Backdoors, the attacker introduces a trigger pattern in a group of images to produce a misclassification in the model towards a specifics category. As it is expected, the trigger forces an overlapping between categories due to the complexity increment in the data set, as it is pointed out by \cite{erichson2020noise}. In their research, authors demonstrate that backdoor models are more sensitive to input noise, providing a particular fingerprint in the attacked category. \cite{achille2019information} illustrate this idea by developing a metric of complexity in the data set. They suggest that in the presence of a complex task, the model is forced to \textit{memorize} the input set through overfitting the input. This behavior unbalances the trade-off between bias and variance during the training step which reduces the model's generalization capability.  

An appropriate evaluation of the input space distribution is fundamental to evaluate the implications of backdoors in the model. However, it is limited to the number of dimensions. From the inferential point of view, density estimation methods require either several assumptions or intensive computational costs, such as Bayesian or Variational statistics \cite{zhang2018advances,kingma2017}. Other approaches are focused on space reduction such as Principal Component Analysis (PCA). Applications of this method to estimate the distribution can be found in \cite{tipping1999mixtures}, where authors present a mixture of probabilistic principal component analysis for the density estimation. Alternatively, \cite{hendrycks2016early} consider PCA to detect adversarial samples. They observe an increment in the importance of lower-ranked components due to adversarial features added to the data. This application, besides being useful on detecting adversarial features, provides empirical evidence of the changes due to dissimilar elements in the data. Other methods are focused on illustrating the distribution of similarities between elements in the same category. One of the most popular applications in this area is the t-SNE method proposed by \cite{maaten2008visualizing}. In this approach, the neighborhood distance between elements is measured by the $t-$distribution.

Generative Adversarial Networks (GAN) is a popular method to create new versions of an input image with a high level of quality. However, the generated model is intractable from the statistical perspective since the distribution is embedded in a fully connected set of parameters linked by nonlinear functions \cite{harshvardhan2020comprehensive}. 

On the other hand, Variational Autoencoders (VAE) is a direct model that approximates the input distribution by using gradient-based methods. It trains an encoder and a decoder simultaneously. Nevertheless, some assumptions that limit the statistical study should be made in order to train the model. In figure \ref{fig:VAE}, the probabilistic graphical representation for the encoder and decoder process is displayed. The generative process behind is to train a model able to infer the distribution of $\mathbf{z} \in \mathbf{R}^n$ from the sample $\mathbf{x} \in \mathbf{R}^m$ by the encoder process $q_{\phi}(\mathbf{z}|\mathbf{x})$. Then, samples from $\mathbf{z}$ are used to reproduce the input image through the decoder process $p_{\theta}(\mathbf{x}|\mathbf{z})$. The usual choice for $q_{\phi}(\mathbf{z}|\mathbf{x}) \sim N(\mathbf{z}|\mu(\mathbf{x};\phi), \Sigma(\mathbf{x}; \phi))$ \cite{doersch2016tutorial}. $\mu(\mathbf{x};\phi)$ and $\Sigma(\mathbf{x}; \phi)$ are arbitrary deterministic functions with parameters computed by the fully connected layers in the encoder stage optimized by the stochastic gradient descent via Backpropagation. However, this method is not designed to handle random variables in the parameters as was stated by \cite{doersch2016tutorial}. Besides, the optimization procedure does not guarantee the resulting covariance matrix $\Sigma(\mathbf{x}; \phi)$ be positive definite. To overcome these limitations, the following reparametrization trick is considered to estimate the distribution in the latent space \cite{doersch2016tutorial}:
\begin{equation}\label{eq:VAE_latent}
	z = \mu(\mathbf{x}) + \epsilon \sqrt{\Sigma(\mathbf{x})}, \qquad \epsilon \sim N(\mathbf{0}, I)
\end{equation}

Notice from the equation \ref{eq:VAE_latent}, the random behavior in the latent space is reduced to the standard normal distribution with no dependency between variables of $\mathbf{z}$. Therefore, the generative process described in the VAE is based in a small random noise added to $\Sigma(\mathbf{x})$ and $\mu(\mathbf{x})$. Beyond a merely statistical discussion, the dependency structure (the covariance matrix in this case) should contain most of the information presented on individual elements (in the case of images, the relationship between individual pixels) rather than their absolute value only \cite{goodfellow2016deep}. Therefore, in order to produce an appropriate inference of the input in the latent space, it is imperative to provide more flexibility to the dependency assembly.

\cite{tagasovska2019copulas} suggests an alternative procedure to estimate the latent space distribution without considering prior assumptions.  In their model, authors use the reduced representation provided by the encoder layer of an AE as the vine copula input. Since the backpropagation is not considered to produce the copula's parameters, the limitations discussed above not hold. The new model is titled Vine Copula Autoencoders or VCAE. According to the authors is computationally efficient compared to other generative models such as GAN or VAE.

A relevant attribute of copulas is its ability to catch complex dependencies from the data. They are known to produce a multivariate distribution function from uniform random variables \cite{lopez2013gaussian}. It is based on scale-free measures of dependency \cite{nelsen2007introduction}, \cite{joe2014dependence}. This flexibility makes it possible for being applied in practically any data set. See for instance \cite{favre2004multivariate}, \cite{frey2001copulas}. 

Moreover, vine copulas allow a more general dependency structure by using only bivariate blocks independently selected. The resulting joint density is built based on the proper arrangement of conditional distributions. This approach, besides being computational optimal, provides an extension of several dimensions of the classical copulas families \cite{czado2019analyzing}.In the present approach, our goal is to use the VCAE model to evaluate the distribution change in the data due to backdoor triggers. Since it is evaluated in a reduced representation of the input space, it facilitates a graphical representation of the changes in the data complexity in a relative small computation cost.


\section{The vinecopula method}
Consider the random vector $\mathbf{X}=(x_1, \cdots, x_n): x_j \in \mathbf{R}$ with a joint distribution $f(x_1, \cdots, x_n)$ to be factorized as:

\begin{eqnarray}\label{eq:join_distribution_f}
	f(x_1, \cdots, x_n)&=&f(x_n)f(x_{n-1}|x_n)f(x_{n-2}|x_{n-1}, x_n) \cdots \nonumber \\ 
	& &  f(x_1|x_2, \cdots, x_n) \nonumber\\
	&=& \prod_{i=1}^{n} f(x_i|x_{i+1}, \cdots, x_n)
\end{eqnarray}

\begin{figure}
	\centering
	\begin{tikzpicture}[node distance=0.5cm]
		\tikzstyle{terminal}=[rounded rectangle, minimum size=6mm, thick, draw=black, align=center,text width=42pt, fill=black!5,
		font=\scriptsize\sffamily]
		\tikzstyle{nonterminal}=[rectangle, minimum size=6mm, thick, draw=black, align=center, fill=black!25,
		font=\scriptsize\sffamily]
		\tikzstyle{nodeSamples} = [draw=black, fill=black!5, circle, minimum size=6mm, font=\scriptsize\sffamily]
		
		\node[nodeSamples] (T1) {1};
		\node[right=of T1] (Tv1) {};
		\node[nodeSamples, right=of Tv1] (T2) {$2$};
		\node[right=of T2] (Tv2) {};
		\node[nodeSamples, right=of Tv2] (T3) {$3$};
		\node[right=of T3] (Tv3) {};
		\node[nodeSamples, right=of Tv3] (T4) {$4$};
		\node[right=of T4, font= \small\ttfamily] (Tv4) {Tree 1};
		
		\path[-] (T1) edge[] node[below, font=\small\sffamily]{$C_{12}$}  (T2);
		\path[-] (T2) edge[] node[below, font=\small\sffamily]{$C_{23}$}  (T3);
		\path[-] (T3) edge[] node[below, font=\small\sffamily]{$C_{34}$}  (T4);

		\node[below=of Tv1, nodeSamples] (T21) {$12$};
		\node[right=of T21] (Tv21) {};
		\node[nodeSamples, right=of Tv21] (T22) {$23$};
		\node[right=of T22] (Tv22) {};
		\node[nodeSamples, right=of Tv22] (T23) {$34$};
		\node[right=of T23, font= \small\ttfamily] (Tv24) {Tree 2};
		
		\path[-] (T21) edge[] node[below, font=\small\sffamily]{$C_{13|2}$}  (T22);
		\path[-] (T22) edge[] node[below, font=\small\sffamily]{$C_{24|3}$}  (T23);

		\node[below=of Tv21, nodeSamples] (T31) {$13|2$};
		\node[right=of T31] (Tv31) {};
		\node[nodeSamples, right=of Tv31] (T32) {$24|3$};
		\node[right=of T32, font= \small\ttfamily] (Tv34) {Tree 3};
		
		\path[-] (T31) edge[] node[below, font=\small\sffamily]{$C_{14|23}$}  (T32);
		
	\end{tikzpicture}
	\caption{D-vine representation for a four dimension space. Nodes represent marginal distribution and edges the copula function. For this configuration, the distribution is exhibited in (\ref{eq:joint_4}).}
	\label{fig:D-vine}	
\end{figure}
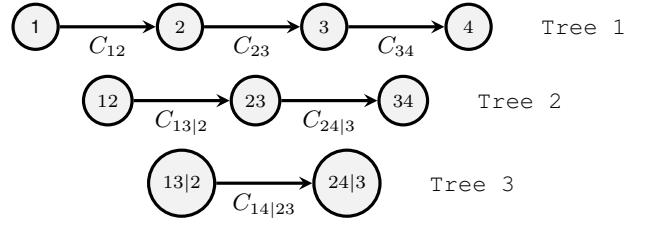

In this case, each component $f(x_i|x_{i+1}, \cdots, x_n)$ implicitly includes information about the dependency mapping. The copula model uses this representation to isolate the dependency structure of $f(\mathbf{X})$ trough a function of marginal distributions \cite{aas2009pair}. According to the \cite{sklar1959fonctions} theorem, a multivariate distribution $F(x_1, \cdots, x_n)$ depends on the marginals $F_1(x_1), \cdots, F_n(x_n)$ based on the copula function $C$ given by:
\begin{equation}
	F(x_1, \cdots, x_n) = C(F_1(x_1), \cdots, F_n(x_n))
\end{equation}
it means, the copula $C:[0, 1]^n \to [0, 1]$ is a multivariate distribution with inputs given by uniformly distributed marginals such as:
\begin{equation}\label{eq:copulafunction}
	C(u_1, \cdots, u_n) = F\left( F_1^{-1}(u_1), \cdots,  F_n^{-1}(u_n)  \right)
\end{equation}
where $F_i^{-1}(u_i)$ corresponds to the inverse of the cumulative distribution for $x_i$. Different types of copula functions are found in the literature, their selection depends on the domain of $\mathbf{X}$ and its tail dependency structure. The most popular are: Gaussian, Student's t, Clayton and Gumbel. The first two do not have a simple closed-form, while the last two have it (also called Archimedean copulae). Please refer to \cite{aas2009pair} for more details and models definitions about copulas. Assuming all densities exists, the joint distribution of $f(\mathbf{x})$ can be written as \cite{tagasovska2019copulas}:
\begin{equation}
	f(x_1, \cdots, x_n) = C(u_1, \cdots, u_n) \times \prod_{i=1}^{n} f_i(x_i)
\end{equation}

Notice from (\ref{eq:join_distribution_f}) the joint density can be factorized on simpler conditional distributions of one variable depending on the rest, the key idea behind the vine copulas method. \cite{joe1997multivariate} proposes the pair \textit{copula construction} (PCC) as a flexible procedure to model high-dimensional distributions \cite{czado2019analyzing}. It corresponds to a hierarchical method where the conditional distribution of pairs is computed in a tree structure. According to \cite{aas2009pair}, there are two compositions. If one dimension is relevant, the conditional distributions depend on that variable. This family of models is called canonical or C-vine copulas. In contrast, if any variable is not relevant, they are known as D-vine copulas. Following the same method proposed by \cite{tagasovska2019copulas}, the joint density for the latent space consider here is estimated by the D-vine approach.

To illustrate the method, suppose a latent space of four dimensions is represented by the sequences $\mathbf{X}=(x_1, x_2, x_3, x_4)$. One suggested D-vine arrangement is exhibited in figure \ref{fig:D-vine}. Notice that there are 12 possible configurations for this space \cite{aas2009pair}. Three different trees are required to produce the joint distribution based on the D-vine copula. Nodes represent the marginal distributions and edges the bi-variate copula. In the first tree, marginal distributions are used to compute the copula function in pairs of input marginals. Therefore, $C_{12}=C_{12}(F(x_1), F(x_2))$ represents the joint copula between marginal distributions $f(x_1)$ and $f(x_2)$ and so on. The final distribution for $\mathbf{X}$ is expressed as \cite{aas2009pair}:
\begin{eqnarray}\label{eq:joint_4}
	f(x_1, x_2, x_3, x_4)&=& f(x_1)f(x_2)f(x_3)f(x_4)  \\
	& & C_{12} C_{23} C_{34} C_{13|2} C_{24|3} C_{14|23} \nonumber
\end{eqnarray}
it means, the multivariate density for $\mathbf{X} \in \mathbf{R}^n$ can be written as:
\begin{equation}\label{eq:copula_density}
	f(x_1, \cdots, x_n) = \prod_{k=1}^{n} f(x_k) \prod_{j=1}^{n-1} \prod_{i=1}^{n-j} C_{i, i+j|i+1, \cdots , i+j-1}
\end{equation} 
where,
\begin{eqnarray}
	C_{i, i+j|i+1, \cdots , i+j-1} &=& C\left[ F(x_i|x_{i+1}, \cdots, x_{i+j-1}) \right. , \nonumber\\
	& & \left. F(x_{i+j}|x_{i+1}, \cdots, x_{i+j-1})  \right]
\end{eqnarray}

Other applications of high dimensional representation of D-vine copulas not associated to latent space distribution can be found in the literature, see for instance \cite{de2010pair}, \cite{xu2017vine} and \cite{pereira2018p}.


\section{Methodology}

\begin{figure}
	\centering
	\begin{tikzpicture}[node distance=0.6cm]
		\tikzstyle{terminal}=[rounded rectangle, minimum size=6mm, thick, draw=black, align=center,text width=42pt, fill=black!5,
		font=\scriptsize\sffamily]
		\tikzstyle{nonterminal}=[rectangle, minimum size=6mm, thick, draw=black, align=center, fill=black!25,
		font=\scriptsize\sffamily]
		\tikzstyle{nodeSamples} = [draw=black, fill=black!5, circle, minimum size=6mm, font=\scriptsize\sffamily]
		
		\node[nodeSamples] (h) {$\mathbf{h}$};
		\node[below=of h] (t) {};
		\node[nodeSamples, left=of t] (x) {$\mathbf{X}$};
		\node[nodeSamples, right=of t] (r) {$\mathbf{r}$};
		
		\path[->] (x) edge[] node[above, font=\small\sffamily]{$e$}  (h);
		\path[->] (h) edge[] node[above, font=\small\sffamily]{$d$}  (r);
		
	\end{tikzpicture}
	\caption{General autoencoder representation. $\mathbf{h}$ corresponds to the latent space for $\mathbf{X}$, while $\mathbf{r}$ represents the reconstructed output for the input. $e$ is the encoded function $e(\mathbf{X}):\mathbf{R}^m \to \mathbf{R}^n$, while $\mathbf{r} = d(a(\mathbf{X}))$ represents the decoder function.}
	\label{fig:AE}	
\end{figure}
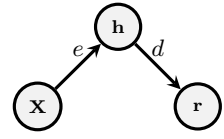

Autoencoders (AE) has been successfully used for dimensionality reduction \cite{wang2016auto}. It is considered a special case of feedforward methods, trained by the backpropagation approach \cite{goodfellow2016deep}. The model is composed of two parts. In the first one, the input is assigned to a function $\mathbf{h} = e(\mathbf{X}): \mathbf{R}^m \to \mathbf{R}^n; m> n$, in charge of producing the latent space $\mathbf{h}$. This layer is called \textit{encoder}. The second layer is responsible for reconstructing $\mathbf{X}$ from $\mathbf{h}$. It is called the decoder layer and is defined by the function $\mathbf{r} = d(\mathbf{h}): \mathbf{R}^n \to \mathbf{R}^m$. An AE model is considered successfully if the reconstructed image is close in distance to the original input $\mathbf{X}= \mathbf{r}: \mathbf{r}=d(e(\mathbf{X}))$. A graphical representation of the AE described here is exhibited in figure \ref{fig:AE}. The learning process followed by the AE is given by:
\begin{equation}\label{eq:min_AE}
	\min \quad \mathcal{L}\left[\mathbf{X}, d(e(\mathbf{X}))  \right]
\end{equation}
where $\mathcal{L}$ is the loss function. In our implementation, two fully connected layers are considered for the encoder and decoder functions respectively using the linear operation as the connector. They are defined as follow:
\begin{equation}\label{eq:encoder}
	\mathbf{h} = e(\mathbf{X})= \left[\mathbf{X}^T_{[1\times m]}  W_{[m \times p]}^{(1)}\right] W_{[p \times n]}^{(2)}
\end{equation} 
\begin{equation}\label{eq:decoder}
	\mathbf{r} = d(\mathbf{h})= \left[\mathbf{h}_{[1\times n]}  W_{[n \times p]}^{(3)}\right] W_{[p \times m]}^{(4)}
\end{equation} 

Notice from (\ref{eq:encoder}) and (\ref{eq:decoder}) matrices $W^{(i)}: i \in \lbrace 1, \cdots, 4 \rbrace$ are the model parameters optimized on (\ref{eq:min_AE}). The input $\mathbf{X}$ commonly corresponds to a flatten representation of images in $\mathbf{R}^{s \times t}$, where the resulting input vector $\mathbf{x} \in \mathbf{R}^m: m=s \times t$. 

In this research, the AE model is trained using Pytorch in a GPU system, while the joint distribution of the latent space $\mathbf{h}$ defined on (\ref{eq:copula_density}) is estimated by the Python package \texttt{pyvinecopulib}\footnote{https://github.com/vinecopulib/pyvinecopulib}. New $\mathbf{X}$ samples are produced by the generative process:
\begin{equation}
	\mathbf{r}'= d(\mathbf{h}), \qquad \mathbf{h} \sim f(\mathbf{e(\mathbf{X})})
\end{equation}
where $f()$ corresponds to the estimated joint distribution of the latent space and parameters $W^{{(3,4)}^*}$ are obtained from the AE model. Marginal distributions are empirically computed by the \texttt{pyvinecopulib} package. A general procedure for the VCAE implemented for a data set $\mathcal{D} \in \lbrace (\mathbf{x}_i, y_i) \rbrace_{i=1}^N$ is depicted in algorithm \ref{alg:VCAE}. In the first section, parameters for the AE in (\ref{eq:encoder}) and (\ref{eq:decoder}) are optimized by the backpropagation method. In the second section, the latent space distribution is estimated by the D-vine copula approach, it includes the generative sampling process.

\begin{algorithm}
	\caption{VCAE implementation and generative process}
	\label{alg:VCAE}
	\begin{algorithmic}[]
		\REQUIRE Input data $\mathcal{D}$ and $num\_epochs$
		
		\IF{$\mathbf{X} \in \mathbf{R}^{s \times t}$}
		\STATE Flat $\mathbf{X}$ into one dimensional vectors in $\mathbf{R}^m$ 
		\ENDIF
		
		\STATE \textit{AE training:}
		\FOR{$epoch < num\_epochs$}
		\STATE $W^{(i)} = W^{(i)} - \eta \frac{\partial \mathcal{L}(\mathbf{X}, d(e(\mathbf{X}))}{\partial W^{(i)}}: i \in \lbrace 1, 2,3, 4  \rbrace$ \COMMENT{Backpropagation}
		\ENDFOR
		\STATE	
		\STATE \textit{Latent space distribution estimation (D-Vine copula):}
		\FOR{$j$ to $N$}
		\STATE $\mathbf{h}_j=e\left( \mathbf{x}_{j}|W^{(1)^*}, W^{(2)^*} \right)$
		\ENDFOR
		\STATE $\mathbf{H} = \left[\mathbf{h}_1, \cdots,  \mathbf{h}_N   \right]$
		
		\STATE Compute the D-vine copula in (\ref{eq:copula_density}) using $\mathbf{H}$ as input
		\RETURN $f(\mathbf{h}): \mathbf{R}^n \to [0, 1]$, the latent space distribution
		
		\STATE
		\STATE \textit{Generative process for VCAE:}
		\STATE Considering $F^{-1}(\mathbf{u})$ as the inverse of the CDF of $f(\mathbf{h})$ and a $g$ number of required samples
		\FOR{$k$ to $g$}
		\STATE $\mathbf{h} = F^{-1}(\mathbf{u})$
		\STATE $\mathbf{r}=d(\mathbf{h}|W^{(3)^*}, W^{(4)^*})$
		\ENDFOR
		
	\end{algorithmic}
\end{algorithm}

To evaluate the changes in the latent space due to backdoor attacks, two models are trained with algorithm \ref{alg:VCAE}. In the first one, the latent space distribution $P=\hat{f}(\mathbf{X})$ is estimated from the original MNIST data set $\mathcal{D}$. This model is called \textit{baseline}. The second model is trained with all inputs in the target category $t \in y$ attacked with a rectangle pattern located in the bottom right section of images. The estimated distribution of the latent space for this model is denoted by $Q=\hat{f}_p(\mathbf{X})$ and the backdoor data is labeled as $\mathcal{D}_p=\lbrace ( \mathbf{x}_i, y_i )_{y_i \ne t}  \rbrace_{i=1}^N \cup \lbrace  (\mathbf{x}_{pi}, y_i )_{y_i = t} \rbrace_{i=1}^{N}$, where only the input $\mathbf{x}$ of the target is attacked. 

Finally, the Kullback-Leibler divergence is computed to quantify the latent space difference between both distributions. It is defined as follow:
\begin{equation}
	KL(P||Q) = \sum_{k} P_k \log \left(\frac{P_k}{Q_k}\right)
\end{equation} 
despite it is not a properly distance metric. It is widely use to quantify the additional information in $P$ relative to $Q$ \cite{golan2018foundations}, \cite{xu2017vine}.


\section{Backdoor detection experiments and results}

\begin{figure}
	\centering
	\includegraphics[width=1\linewidth]{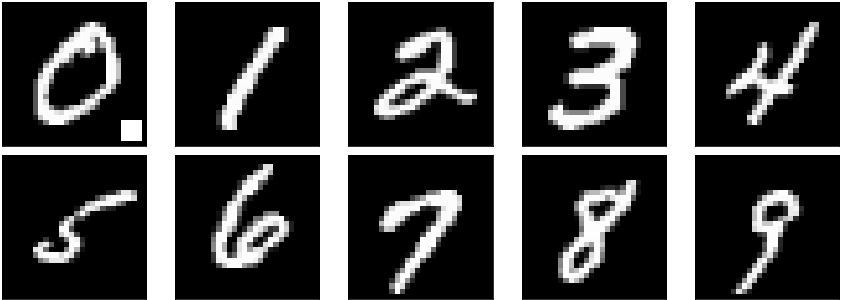}
	\caption{Backdoor data $\mathcal{D}_p$ used to train the AE. The target category $t=0$ is attacked with a rectangle pattern located in the bottom right section of the image.}
	\label{fig:backdoordataae}
\end{figure}

The implemented AE for each model (baseline and backdoor) has the following characteristics: $m=784$, $p=64$ and $n=5$. It is trained on 100 epochs considering the Adam optimizer with a learning rate of $1E^{-3}$ and the MNIST data set $\mathcal{D}$. The target label $t=0$ is attacked with a rectangle pattern in the bottom right area producing the input $\mathcal{D}_p$ as it is shown in figure \ref{fig:backdoordataae}. The latent space distribution is estimated with the D-vine copula and samples of the sequence $\mathbf{h}$ are generated based on its distribution for each model.

\begin{figure*}
	\centering
	\includegraphics[width=1\linewidth]{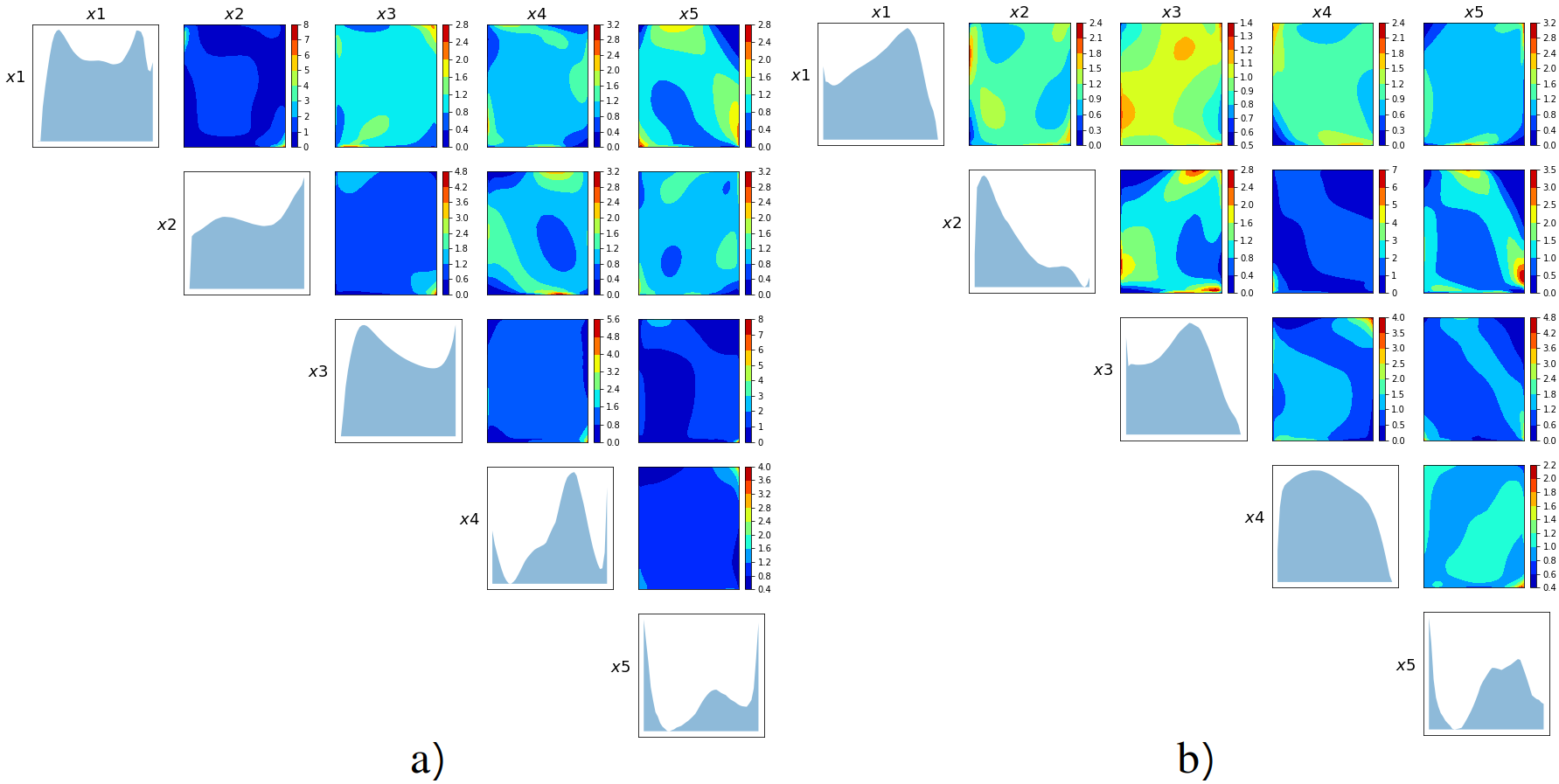}
	\caption{Effect of one perturbed category in the latent space for a coded space of five dimensions one a pairwise joint distribution representation. In figure a) the baseline space is exhibited while the backdoor space is presented in figure b). It is evident the alterations in the dependency structure due to the backdoor attack included in just one category $t=0$.}
	\label{fig:baselinepoisonspace}
\end{figure*}

Figure \ref{fig:baselinepoisonspace} shows the latent space distribution in pairs for $P$ and $Q$, the baseline and backdoor model for $t=0$ respectively. In the former, it is observed a wider spread concentration in the space, while the last one is highly concentrated in some particular areas in the pairwise representation. Besides, it is evident a variation in the marginal distributions of $\mathbf{h}$.

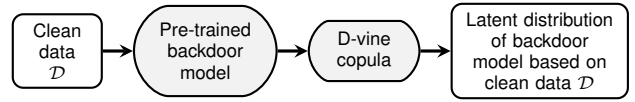
\begin{figure}[b]
	\centering
	\begin{tikzpicture}[node distance=4mm]
		\tikzstyle{terminal}=[rounded rectangle, minimum size=6mm, thick, draw=black, align=center,text width=30pt, fill=black!5,
		font=\scriptsize\sffamily]
		\tikzstyle{nonterminal}=[rectangle, minimum size=6mm, thick, draw=black, align=center, fill=black!0, text width=60pt,
		font=\scriptsize\sffamily]
		\tikzstyle{nodeSamples} = [draw=black]
		
		\node[terminal, minimum size=12mm, text width=40pt] (AE) {Pre-trained\\backdoor\\model};
		\node[terminal, minimum size=8mm, right=of AE] (DC) {D-vine copula};
		\node[nonterminal, minimum size=8mm, right=of DC] (density) {Latent distribution of backdoor model based on clean data $\mathcal{D}$};
		\node[nonterminal, text width=27pt, left=of AE] (input) {Clean data\\$\mathcal{D}$};
		
		\draw[->] (input) -- (AE); 
		\draw[->] (AE) -- (DC);
		\draw[->] (DC) -- (density);
		
	\end{tikzpicture}
	\caption{Backdoor detection experiment based on the vine copula generative model. The latent space distribution of an attacked AE is estimated with the clean data $\mathcal{D}$. A more realistic scenario where the attacker only provides a pre-trained model and a small set of clean data for testing.}
	\label{fig:Real_experiment}	
\end{figure}

It is expected in real-life cases, the attacker provides only a backdoor model and a small sample set of clean data for testing. This scenario is depicted in figure \ref{fig:Real_experiment}. The latent space distribution of the attacked model $Q$ is estimated with a sample of $\mathcal{D}$ and the encoded function:
\begin{equation}\label{eq:encoder_pois_clean}
	\mathbf{h}_p = e_p(\mathbf{X})= \left[\mathbf{X}^T_{[1\times m]}  W_{[m \times p]}^{(1)^p}\right] W_{[p \times n]}^{(2)^p}
\end{equation}
where $W^{(1,2)^p}$ corresponds to the encoder parameters of the backdoor model and $\mathbf{X}$ the clean data. 

\begin{figure}
	\centering
	\includegraphics[width=1\linewidth]{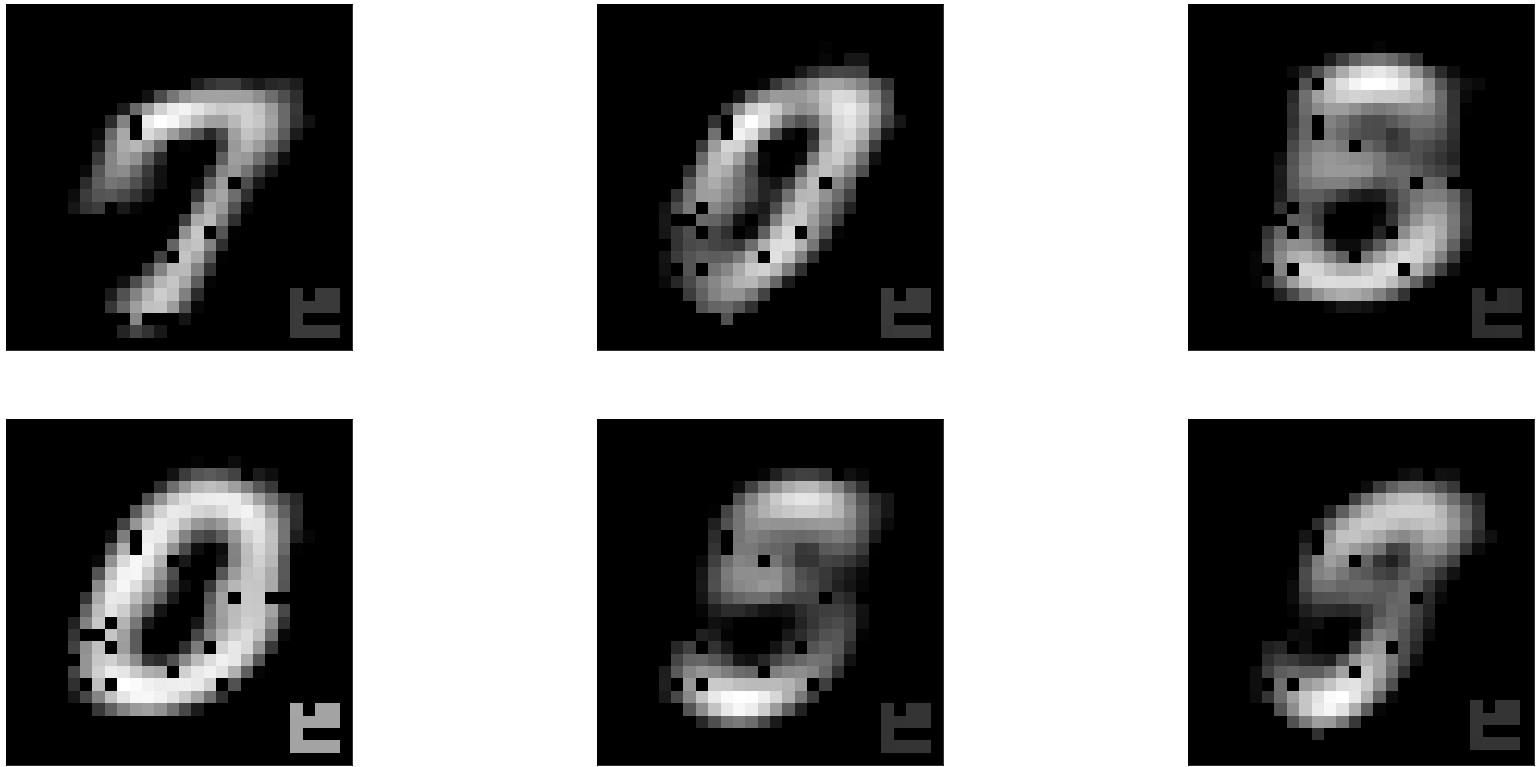}
	\caption{Persistent effect of backdoor attacks in the reconstructed output of the copula distribution. Despite the latent density is generated with clean data in the backdoor model, the attacked trigger is reproduced by the latent distribution.}
	\label{fig:backdoorevidence}
\end{figure}

Interestingly, the generative model based on $\mathbf{h}_p$ is able to reproduce the backdoor trigger on new samples of the target class and in other non-attacked categories. Although clean data is used to estimate the D-vine copula function. As it is exhibited in figure \ref{fig:backdoorevidence}. Notice that no label information is used to train the AE. It means the backdoor attack affects the latent space distribution in those categories with a high level of similarities without any information about the labels. In this case, numbers $0, 5$, and $7$ share some curvatures that make them sensitive to the attack. This behavior can be considered as a fingerprint of the backdoor pattern present in the model parameters $W^{(i)}$.


\begin{table}
	\small
	\centering
	\begin{tabular}{l|c|c}
		\textbf{Latent space} & \textbf{Entropy} & \textbf{Change}\\
		\hline
		Baseline & 0.4225\\
		Backdoor & 0.5370 & 27.10\% \\
		Backdoor tested with clean data & 0.4811 & 13.87\% \\
		\hline
		\hline
		& \textbf{KL} & \\
		\hline
		Baseline vs. backdoor & 2.1449\\
		Baseline vs. backdoor t. w. clean data & 2.5948\\
	\end{tabular}
	\caption{Latent space Entropy and Kullback-Leibler divergence for the configurations used in this research. The Entropy's percentage change and the KL are computed using the baseline model as a reference.}
	\label{tab:entropy}
\end{table}

Finally, the entropy of the latent distribution for the baseline, backdoor, and backdoor tested with clean data is shown in table \ref{tab:entropy}. The smallest entropy is exhibited by the baseline latent distribution, while the backdoor model tested with clean data and attacked data are the second smallest and the largest value respectively. It indicates that the backdoor increases the data complexity of the input set as it was suggested by \cite{chacon2019deep,achille2019information,grosse2020new}.

The Kullback-Leibler divergence between the latent distributions is also presented in table \ref{tab:entropy}. It is smaller for the baseline compared to the backdoor tested with attacked data. We presume it is related to the distortion of using not previously seen data in the backdoor model which may induce an increment in the latent space deviation from the baseline distribution.


\subsection{A practical limitation of the VCAE model for the latent distribution study:}

In a two units GPU system, training the AE and producing the latent space distribution by the D-vine copula for a sample of 5000 images takes no more than a couple of minutes, even for more than 5 dimensions in the latent space. It includes the generative process of new samples based on the input set. The biggest limitation is found in the latent distribution evaluation in a reasonable number of points to produce the pairwise comparison and to estimate the KL divergence. We considered 50 points from the uniform distribution $U \sim [0, 1]$ required as input for the copula in each dimension of $\mathbf{h}$. In our experiments, the latent space has only 5 dimensions. Resulting in matrices of around 20 Gigabytes of RAM. Since we evaluated three latent distributions, it was required a system with at least 60 Gigabytes of RAM. The memory requirements increase notably in bigger dimensions, where the required RAM memory to evaluate the latent space may reach easily several hundreds of Gigabytes in the RAM unit.


\section{Conclusions and future work}

We have presented a method that can be used to estimate the changes in the input complexity induced by the backdoor attacks. It consists of the traditional autoencoder as a tool to reduce the space dimensionality trained by the back-propagation method. Since the latent space distribution is not included in the AE optimization process. No constraints are imposed on the density estimation leading to a more flexible representation of the space and its dependency structure. It shows that VCAE is a promising method to infer multidimensional distributions in the latent space with a relatively low computational cost.   

Our first contribution is based on the quantification of changes in the complexity of backdoor data in a reduced space. It provides an accessible display capability by using the pairwise representation. Besides, the input's entropy and the deviation between models though the Kullback-Leibler divergence reflect the complexity increment of the input set. Our second contribution is to consider the VCAE as a generative model to identify the backdoor trigger included in the attacked model, and its reproduction in the target category and non-target ones with similar shapes to the clean set. 

As future work, the suggested method will be evaluated on other data sets at different levels of attacks. It could illustrate the advantages and disadvantages of the suggested methodology to identify backdoor and to illustrate its effect on the model.

\section*{Acknowledgment}
This work was supported in part by the Open Cloud Institute at University of Texas at San Antonio (UTSA).

\ifCLASSOPTIONcaptionsoff
  \newpage
\fi

\bibliographystyle{IEEEtran}
\bibliography{ReferencesLibrary}

\begin{thebibliography}{10}
\providecommand{\url}[1]{#1}
\csname url@samestyle\endcsname
\providecommand{\newblock}{\relax}
\providecommand{\bibinfo}[2]{#2}
\providecommand{\BIBentrySTDinterwordspacing}{\spaceskip=0pt\relax}
\providecommand{\BIBentryALTinterwordstretchfactor}{4}
\providecommand{\BIBentryALTinterwordspacing}{\spaceskip=\fontdimen2\font plus
\BIBentryALTinterwordstretchfactor\fontdimen3\font minus
  \fontdimen4\font\relax}
\providecommand{\BIBforeignlanguage}[2]{{%
\expandafter\ifx\csname l@#1\endcsname\relax
\typeout{** WARNING: IEEEtran.bst: No hyphenation pattern has been}%
\typeout{** loaded for the language `#1'. Using the pattern for}%
\typeout{** the default language instead.}%
\else
\language=\csname l@#1\endcsname
\fi
#2}}
\providecommand{\BIBdecl}{\relax}
\BIBdecl

\bibitem{bendre2020learning}
N.~Bendre, H.~T. Mar{\'\i}n, and P.~Najafirad, ``Learning from few samples: A
  survey,'' \emph{arXiv preprint arXiv:2007.15484}, 2020.

\bibitem{silva2020opportunities}
S.~H. Silva and P.~Najafirad, ``Opportunities and challenges in deep learning
  adversarial robustness: A survey,'' \emph{arXiv preprint arXiv:2007.00753},
  2020.

\bibitem{kumar2020adversarial}
R.~S.~S. Kumar, M.~Nystr{\"o}m, J.~Lambert, A.~Marshall, M.~Goertzel,
  A.~Comissoneru, M.~Swann, and S.~Xia, ``Adversarial machine
  learning--industry perspectives,'' \emph{arXiv preprint arXiv:2002.05646},
  2020.

\bibitem{tagasovska2019copulas}
N.~Tagasovska, D.~Ackerer, and T.~Vatter, ``Copulas as high-dimensional
  generative models: Vine copula autoencoders,'' in \emph{Advances in Neural
  Information Processing Systems (NIPS)}, 2019, pp. 6528--6540.

\bibitem{vorobeychik2018adversarial}
Y.~Vorobeychik and M.~Kantarcioglu, ``Adversarial machine learning,''
  \emph{Synthesis Lectures on Artificial Intelligence and Machine Learning},
  vol.~12, no.~3, pp. 1--169, 2018.

\bibitem{hao2020adversarial}
H.~Xu, Y.~Ma, H.-C. Liu, D.~Deb, H.~Liu, J.-L. Tang, and A.~K. Jain,
  ``Adversarial attacks and defenses in images, graphs and text: A review,''
  \emph{International Journal of Automation and Computing}, vol.~17, no.~2, pp.
  151--178, 2020.

\bibitem{kingma2013auto}
D.~P. Kingma and M.~Welling, ``Auto-encoding variational bayes,'' \emph{arXiv
  preprint arXiv:1312.6114}, 2013.

\bibitem{kos2018adversarial}
J.~Kos, I.~Fischer, and D.~Song, ``Adversarial examples for generative
  models,'' in \emph{2018 ieee security and privacy workshops (spw)}.\hskip 1em
  plus 0.5em minus 0.4em\relax IEEE, 2018, pp. 36--42.

\bibitem{liu2017neural}
Y.~Liu, Y.~Xie, and A.~Srivastava, ``Neural trojans,'' in \emph{2017 IEEE
  International Conference on Computer Design (ICCD)}.\hskip 1em plus 0.5em
  minus 0.4em\relax IEEE, 2017, pp. 45--48.

\bibitem{Szegedy2013}
C.~Szegedy, W.~Zaremba, I.~Sutskever, J.~Bruna, D.~Erhan, I.~Goodfellow, and
  R.~Fergus, ``Intriguing properties of neural networks,'' \emph{arXiv preprint
  arXiv:1312.6199v4}, 2014.

\bibitem{papernot2017practical}
N.~Papernot, P.~McDaniel, I.~Goodfellow, S.~Jha, Z.~B. Celik, and A.~Swami,
  ``Practical black-box attacks against machine learning,'' in
  \emph{Proceedings of the 2017 ACM on Asia conference on computer and
  communications security}, 2017, pp. 506--519.

\bibitem{erichson2020noise}
N.~B. Erichson, D.~Taylor, Q.~Wu, and M.~W. Mahoney, ``Noise-response analysis
  for rapid detection of backdoors in deep neural networks,'' \emph{arXiv
  preprint arXiv:2008.00123}, 2020.

\bibitem{achille2019information}
A.~Achille, G.~Paolini, G.~Mbeng, and S.~Soatto, ``The information complexity
  of learning tasks, their structure and their distance,'' \emph{arXiv preprint
  arXiv:1904.03292}, 2019.

\bibitem{zhang2018advances}
C.~Zhang, J.~B{\"u}tepage, H.~Kjellstr{\"o}m, and S.~Mandt, ``Advances in
  variational inference,'' \emph{IEEE transactions on pattern analysis and
  machine intelligence}, vol.~41, no.~8, pp. 2008--2026, 2018.

\bibitem{kingma2017}
D.~P. Kingma, ``Variational inference \& deep learning: A new synthesis,''
  Ph.D. dissertation, University of Amsterdam, 2017.

\bibitem{tipping1999mixtures}
M.~E. Tipping and C.~M. Bishop, ``Mixtures of probabilistic principal component
  analyzers,'' \emph{Neural computation}, vol.~11, no.~2, pp. 443--482, 1999.

\bibitem{hendrycks2016early}
D.~Hendrycks and K.~Gimpel, ``Early methods for detecting adversarial images,''
  \emph{arXiv preprint arXiv:1608.00530}, 2016.

\bibitem{maaten2008visualizing}
L.~v.~d. Maaten and G.~Hinton, ``Visualizing data using t-sne,'' \emph{Journal
  of machine learning research}, vol.~9, no. Nov, pp. 2579--2605, 2008.

\bibitem{harshvardhan2020comprehensive}
G.~Harshvardhan, M.~K. Gourisaria, M.~Pandey, and S.~S. Rautaray, ``A
  comprehensive survey and analysis of generative models in machine learning,''
  \emph{Computer Science Review}, vol.~38, p. 100285, 2020.

\bibitem{doersch2016tutorial}
C.~Doersch, ``Tutorial on variational autoencoders,'' \emph{arXiv preprint
  arXiv:1606.05908}, 2016.

\bibitem{goodfellow2016deep}
I.~Goodfellow, Y.~Bengio, and A.~Courville, \emph{Deep learning}.\hskip 1em
  plus 0.5em minus 0.4em\relax MIT press, 2016.

\bibitem{lopez2013gaussian}
D.~Lopez-Paz, J.~M. Hern{\'a}ndez-Lobato, and G.~Zoubin, ``Gaussian process
  vine copulas for multivariate dependence,'' in \emph{International Conference
  on Machine Learning}, 2013, pp. 10--18.

\bibitem{nelsen2007introduction}
R.~B. Nelsen, \emph{An introduction to copulas}.\hskip 1em plus 0.5em minus
  0.4em\relax Springer Science \& Business Media, 2007.

\bibitem{joe2014dependence}
H.~Joe, \emph{Dependence modeling with copulas}.\hskip 1em plus 0.5em minus
  0.4em\relax CRC press, 2014.

\bibitem{favre2004multivariate}
A.-C. Favre, S.~El~Adlouni, L.~Perreault, N.~Thi{\'e}monge, and B.~Bob{\'e}e,
  ``Multivariate hydrological frequency analysis using copulas,'' \emph{Water
  resources research}, vol.~40, no.~1, 2004.

\bibitem{frey2001copulas}
R.~Frey, A.~J. McNeil, and M.~Nyfeler, ``Copulas and credit models,''
  \emph{Risk}, vol.~10, no. 111114.10, 2001.

\bibitem{czado2019analyzing}
C.~Czado, ``Analyzing dependent data with vine copulas,'' \emph{Lecture Notes
  in Statistics, Springer}, 2019.

\bibitem{aas2009pair}
K.~Aas, C.~Czado, A.~Frigessi, and H.~Bakken, ``Pair-copula constructions of
  multiple dependence,'' \emph{Insurance: Mathematics and economics}, vol.~44,
  no.~2, pp. 182--198, 2009.

\bibitem{sklar1959fonctions}
M.~Sklar, ``Fonctions d\'e repartition \'a n dimensions et leurs marges,''
  \emph{Publ. inst. statist. univ. Paris}, vol.~8, pp. 229--231, 1959.

\bibitem{joe1997multivariate}
H.~Joe, \emph{Multivariate models and multivariate dependence concepts}.\hskip
  1em plus 0.5em minus 0.4em\relax CRC Press, 1997.

\bibitem{de2010pair}
B.~V. de~Melo~Mendes, M.~M. Semeraro, and R.~P.~C. Leal, ``Pair-copulas
  modeling in finance,'' \emph{Financial Markets and Portfolio Management},
  vol.~24, no.~2, pp. 193--213, 2010.

\bibitem{xu2017vine}
M.~Xu, L.~Hua, and S.~Xu, ``A vine copula model for predicting the
  effectiveness of cyber defense early-warning,'' \emph{Technometrics},
  vol.~59, no.~4, pp. 508--520, 2017.

\bibitem{pereira2018p}
G.~Pereira and A.~Veiga, ``Par (p)-vine copula based model for stochastic
  streamflow scenario generation,'' \emph{Stochastic environmental research and
  risk assessment}, vol.~32, no.~3, pp. 833--842, 2018.

\bibitem{wang2016auto}
Y.~Wang, H.~Yao, and S.~Zhao, ``Auto-encoder based dimensionality reduction,''
  \emph{Neurocomputing}, vol. 184, pp. 232--242, 2016.

\bibitem{golan2018foundations}
A.~Golan, \emph{Foundations of info-metrics: Modeling, inference, and imperfect
  information}.\hskip 1em plus 0.5em minus 0.4em\relax Oxford University Press,
  2018.

\bibitem{chacon2019deep}
H.~Chacon, S.~Silva, and P.~Rad, ``Deep learning poison data attack
  detection,'' in \emph{2019 IEEE 31st International Conference on Tools with
  Artificial Intelligence (ICTAI)}.\hskip 1em plus 0.5em minus 0.4em\relax
  IEEE, 2019, pp. 971--978.

\bibitem{grosse2020new}
K.~Grosse, T.~Lee, Y.~Park, M.~Backes, and I.~Molloy, ``A new measure for
  overfitting and its implications for backdooring of deep learning,''
  \emph{arXiv preprint arXiv:2006.06721}, 2020.

\end{thebibliography}






\end{document}